\documentclass[10pt,twocolumn,letterpaper]{article}

\usepackage[pagenumbers]{iccv} 

\definecolor{iccvblue}{rgb}{0.21,0.49,0.74}
\usepackage[pagebackref,breaklinks,colorlinks,allcolors=iccvblue]{hyperref}
\usepackage{url}
\usepackage{graphicx}
\usepackage{wrapfig}
\usepackage{caption}
\usepackage{amsmath}
\usepackage{booktabs}
\usepackage{multirow}
\usepackage{tabularx}
\usepackage{float}
\usepackage{array}
\usepackage{placeins}

\title{\textbf{Phantom}: Subject-Consistent Video Generation via Cross-Modal Alignment}

\author{
	Lijie Liu$^{*}$ 
	\hspace{9pt} Tianxiang Ma$^{*}$ 
	\hspace{9pt} Bingchuan Li$^{*\dagger}$
	\hspace{9pt} Zhuowei Chen$^{*}$ 
	\hspace{9pt} Jiawei Liu \\
	\hspace{9pt} Gen Li
	\hspace{9pt} Siyu Zhou
	\hspace{9pt} Qian He
	\vspace{5pt} \hspace{9pt} Xinglong Wu \\
	\large$~$Intelligent Creation Team, $~$ByteDance \hspace{5pt}\\
	\large\url{https://phantom-video.github.io/Phantom/}
}

\begin{document}
\maketitle
\renewcommand\thefootnote{\fnsymbol{footnote}}
\footnotetext{~$^*$Equal Contributions}
\footnotetext{~$^\dagger$Project Lead}
\begin{abstract}
The continuous development of foundational models for video generation is evolving into various applications, with subject-consistent video generation still in the exploratory stage. 
We refer to this as Subject-to-Video, which extracts subject elements from reference images and generates subject-consistent videos following textual instructions. 
We believe that the essence of subject-to-video lies in balancing the dual-modal prompts of text and image, thereby deeply and simultaneously aligning both text and visual content. 
To this end, we propose \textbf{Phantom}, a unified video generation framework for both single- and multi-subject references.
Building on existing text-to-video and image-to-video architectures, we redesign the joint text-image injection model and drive it to learn cross-modal alignment via text-image-video triplet data. 
The proposed method achieves high-fidelity subject-consistent video generation while addressing issues of image content leakage and multi-subject confusion.
Evaluation results indicate that our method outperforms other state-of-the-art closed-source commercial solutions.
In particular, we emphasize subject consistency in human generation, covering existing ID-preserving video generation while offering enhanced advantages.
\end{abstract}    
\section{Introduction}
\label{sec:intro}

The rise of diffusion models \cite{peebles2023scalable, ho2020denoising} is rapidly reshaping the field of generative modeling at an astonishing pace. Among them, the advancements in video generation brought by diffusion models are particularly remarkable. 
In the visual domain, video generation requires to pay more attention to the continuity and consistency of multiple frames compared to image generation, which poses additional challenges. 
Inspired by the scaling laws of large language models \cite{GPT4, touvron2023llama, yang2024qwen2}, the focus of video generation has shifted towards investigating foundational large models, similar to Sora \cite{OpenAI2023Sora, polyak2024movie, yang2024cogvideox, kong2024hunyuanvideo, zheng2024open}, which have demonstrated promising visual effects and are paving the way for a new era in Artificial Intelligence Generated Content. 


\begin{figure}[t]
	\centering
	\includegraphics[width=0.82\columnwidth]{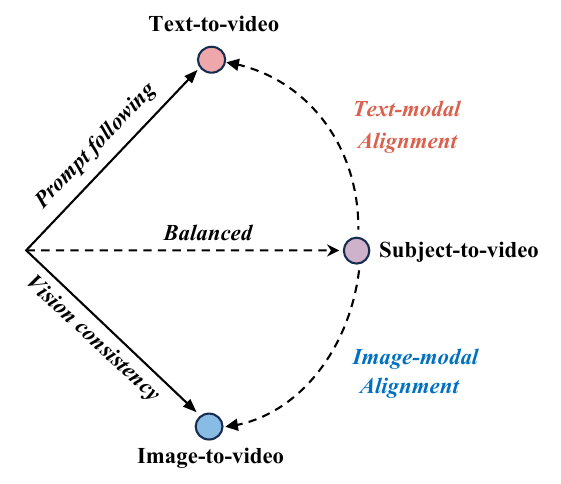}
	\caption{
		Relationship in cross-modal video generation tasks.
	}
	\label{fig:motivation}
\end{figure}

\begin{figure*}[h]
	\centering
	\includegraphics[width=0.98\textwidth]{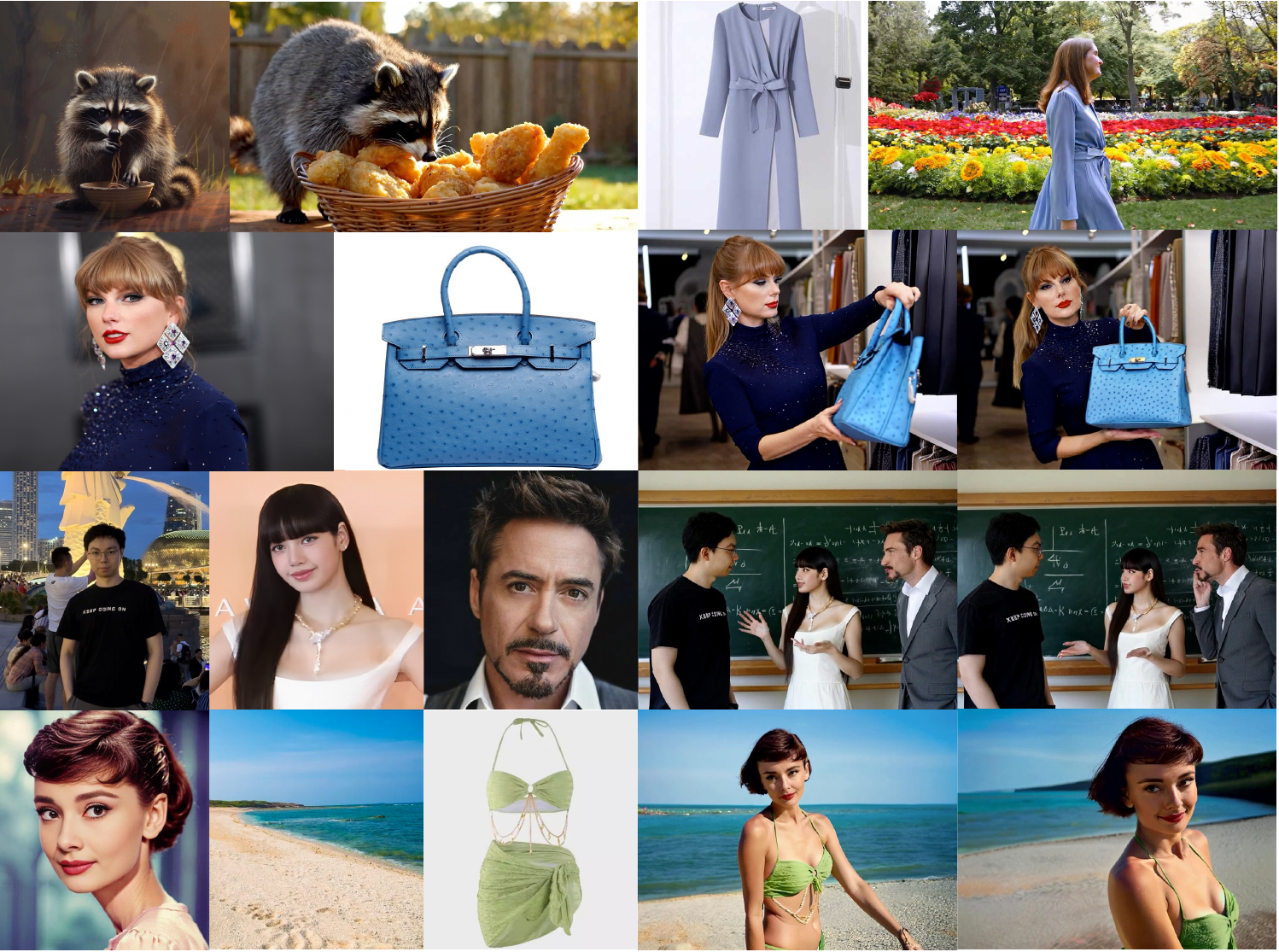}
	\caption{
		Subject-consistent video generation examples using our method, with reference images and corresponding generated video frames (text prompts omitted). The last three rows show multiple reference subjects.}
	\label{fig:teaser}
\end{figure*}

Currently, foundational video generation models focus mainly on two major tasks: text-to-video\cite{OpenAI2023Sora} and image-to-video \cite{blattmann2023stable}. 
Text-to-video (T2V) leverages language models to understand input text instructions and generate visual content describing the expected characters, movements, and backgrounds. While it allows for creative and imaginative content combinations, it often struggles with generating consistently predictable results due to inherent randomness. 
On the other hand, image-to-video (I2V) typically provides the first frame of an image along with optional text descriptions to transform a static image into a dynamic video. Although it is more controllable, the content richness is often limited by the strict ``copy-paste" \cite{polyak2024movie, chen2025multi} nature of the first frame.
We term the process of subject-consistent video generation as subject-to-video (S2V) \cite{huang2025conceptmaster, chen2025multi, Hailuo}, which involves capturing the subject from an image and flexibly generating a video based on text prompts, while combining the diversity and controllability of joint image and text inputs. As shown in Figure \ref{fig:motivation}, its essence lies in balancing the dual-modal prompts of text and image, requiring the model to simultaneously align text instructions and image contents.


However, the research on subject consistency in video generation tasks still lags behind image generation scenarios. 
As text-to-image (T2I) foundation models \cite{flux2024, esser2024scaling} have matured, subject-to-image (S2I) has evolved from parameter optimization methods \cite{ huang2024context, ruiz2023dreambooth} to adapter-based training approaches \cite{ye2023ip, huang2024realcustom}, to unified image editing approaches \cite{chen2024unireal, xiao2024omnigen, wanx_ace}, achieving impressive results (refer to Sec \ref{sec:related_work:image}).
The most straightforward way to implement S2V is to combine S2I with I2V, but there are two main limitations. First, S2I has greater difficulty in learning subject consistency compared to S2V, as the S2V training data naturally include multi-view dynamic variations, allowing for better understanding of the subject.
Second, transitioning from S2I to I2V can lead to information loss. For instance, when generating a back-to-front view motion, the subject's ID information may be lost because the first frame lacks it, which hinders I2V from maintaining ID consistency (see supplementary material).
Therefore, subject-consistent generation requires a specialized video model for unified processing.

\begin{figure*}[t]
	\centering
	\includegraphics[width=\textwidth]{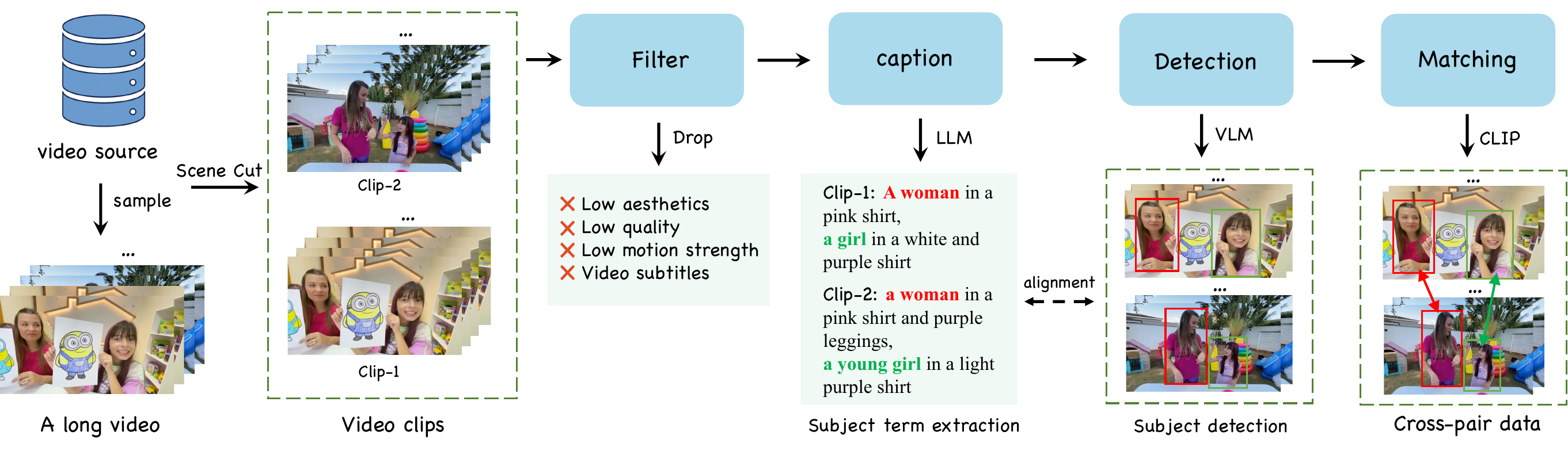} 
	\caption{Data processing pipeline for cross-modal video generation. The process involves filtering, adding captions, detection, and matching stages to extract subjects from video clips and align them with the text prompts, ensuring consistent video generation.}
	\label{fig:data_pipeline}
\end{figure*}

Specifically, the subject-consistent video generation task aims to deeply and simultaneously align the content described in the text and images. To achieve this, we propose a data pipeline for the S2V task, producing training data in the form of text-image-video triplets. Two key issues must be addressed. First, prevent the leakage of image content into the generated video. Some methods \cite{chen2025multi, huang2025conceptmaster, liang2025movie, yuan2024identity, he2024id} sample key frames from a video as image conditions to reconstruct the video, but this allows the model to copy-paste image content, reducing text responsiveness. While some approaches enhance data through transformations \cite{chen2025multi}, they fail to address the rigid properties and overall lighting of the image. We focus on constructing cross-video multi-subject pairs to ensure that subjects exhibit non-rigid deformations and color variations while maintaining content matching. Second, address the issue of confusion arising from multi-subject generation. Specifically, when similar subjects trigger identical textual descriptions, it can lead to content ambiguity. To resolve this, we emphasized distinct descriptions of the subjects' appearances in the video. The appearances of multiple subjects should be distinguishable and precisely match the contents of the sampled reference images. Furthermore, we build a rephraser that rephrases user's input text prompts to include detailed description of the image content.

Our model design is based on two primary considerations: \textbf{(1)} How to simply and effectively extend a video foundation model to support S2V capabilities; \textbf{(2)} How to achieve a unified framework for single- and multi-subject consistency generation. 
Thus, we redesigned the image-text joint injection model based on pre-trained T2V and I2V models \cite{lin2025diffusion} to ensure effective cross-modal learning. 
Specifically, our method is built on the MMDiT \cite{esser2024scaling} architecture. Referring to \cite{wang2025seedvr}, full self-attention is replaced with window attention to reduce computational costs. VAE \cite{esser2021taming} and CLIP \cite{zhai2023sigmoid} are used to encode the reference images, and the encoded results are fed into the video and text branches of MMDiT, respectively.
The VAE latent provides low-level detail information, while CLIP offers high-level semantic information.
Additionally, we introduced a dynamic information injection strategy during attention calculation, allowing the insertion of one or more reference images without affecting the window size and position encoding \cite{su2024roformer}, achieving a unified model architecture for single- and multi-subject consistent video generation.

In addition, for the S2V task, we constructed evaluation datasets for portrait IDs, single subjects, and multiple subjects, and developed corresponding evaluation metrics. Since the performance of some open-source reproducible projects \cite{he2024id, yuan2024identity, huang2025conceptmaster, liang2025movie, chen2025multi} has not yet matched that of closed-source commercial solutions \cite{Vidu, Pika, Keling, Hailuo, polyak2024movie}, our focus is on comparing with commercial methods. Overall, our proposed \textbf{\textit{Phantom}} has the following contributions:

\textbf{Concepts.} (1) We are the first to clearly define the subject-to-video (S2V) task and elucidate its relationship with text-to-video (T2V) and image-to-video (I2V), as in Fig. \ref{fig:motivation}; (2) \textit{Phantom} offers a feasible path for the S2V task, focusing on high-quality alignment of both textual and visual content.

\textbf{Technology.} (1) A new data pipeline constructs cross-modal aligned triplet data, effectively addressing the issues of information leakage (copy-paste) and content confusion (multiple subjects); (2) \textit{Phantom} offers a unified framework for generating videos from both single and multiple subject references, utilizing dynamic injection scheme of various conditions at its core.

\textbf{Significance.} (1) \textit{Phantom} demonstrates superior generation quality, bridging the gap between academic research and proprietary commercialization; (2) the unified consistency generation paradigm covers subtasks such as ID generation and demonstrates significant advantages, indicating that \textit{Phantom}-like solutions have broad prospects in scenarios such as the film industry or advertising production.


\begin{figure*}[t]
	\centering
	\includegraphics[width=0.98\textwidth]{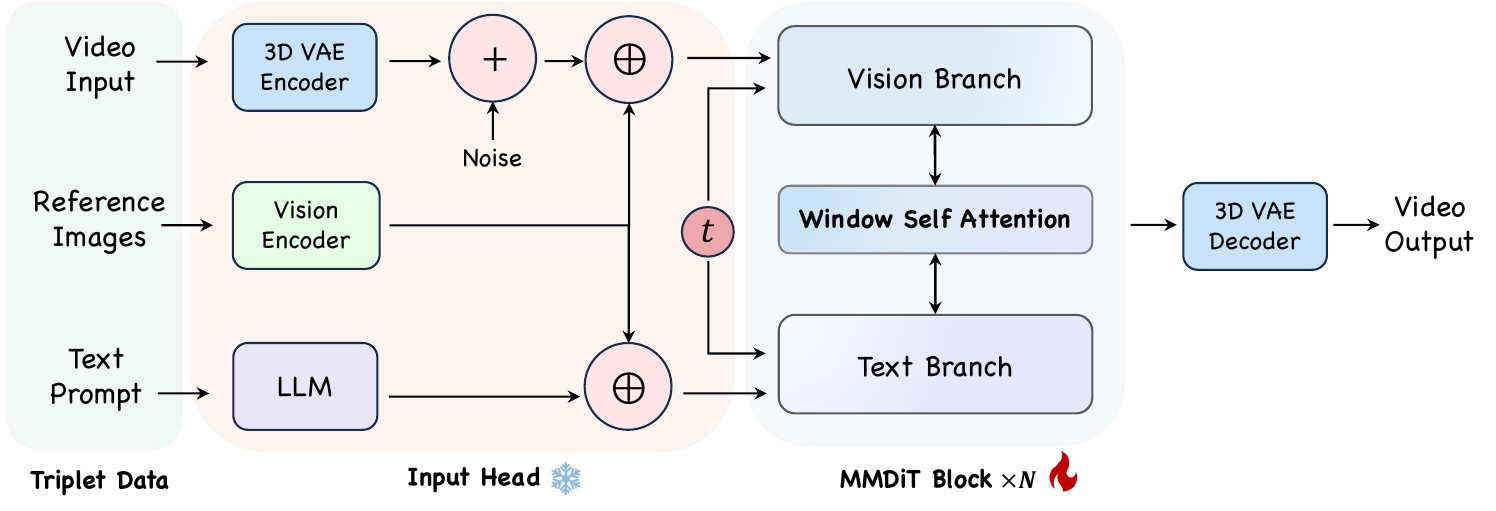} 
	\caption{Overview of the \textbf{\textit{Phantom}} architecture. Triplet data is encoded into latent space at the input head, and after combination, it is processed through modified MMDiT blocks to learn the alignment of different modalities. }
	\label{fig:framework}
\end{figure*}

\section{Related Work}
\label{sec:related_work}

\subsection{Video foundation model}
\label{sec:related_work:foundation}

The diffusion algorithm has spurred the rise of video foundation model research, significantly impacting content creation and intelligent interaction. Early latent diffusion models (LDM) \cite{rombach2022high} typically utilized U-Net \cite{ronneberger2015u} architectures, such as the open-source Stable Diffusion 1.5 \cite{rombach2022high}.  Temporal modules were later added to these models, evolving them into video generation models like Make-A-Video \cite{singer2022make}, SVD \cite{blattmann2023stable}, and Animatediff \cite{guo2023animatediff}. The DiT \cite{peebles2023scalable} architecture, guided by scaling laws, has led to the development of more vision foundation models. Among these, Stable Diffusion 3 introduced MMDiT \cite{esser2024scaling}, a dual-stream DiT architecture, which has been adopted in open-source video generation projects such as CogvideoX \cite{yang2024cogvideox}, HunyuanVideo \cite{kong2024hunyuanvideo} and SeedVR \cite{wang2025seedvr}.

\subsection{Subject-consistency image generation}
\label{sec:related_work:image}

In recent years, significant progress has been made in subject-consistent generation for image tasks. Optimization-based methods \cite{ruiz2023dreambooth, gal2022image, hu2021lora, shah2024ziplora, huang2024context} training bind image content with special identifiers for text-to-image generation. A notable work in the training and inference paradigm is IP-Adapter \cite{ye2023ip}, which freezes the base model weights while only training additional adapters to achieve subject-consistent generation. This approach is also widely used in tasks requiring facial ID consistency \cite{guo2024pulid, wang2024instantid, chen2024dreamidentity}. However, these solutions often rely on CLIP \cite{cherti2023reproducible} or DINO \cite{oquab2023dinov2} for extracting image semantics, leading to a trade-off between low-level detail reconstruction and flexible text response. Recent advancements have unified image generation and editing tasks \cite{chen2024unireal, xiao2024omnigen, erwold-2024-qwen2vl-flux, wanx_ace}, enabling various types of editing tasks within a single model, including subject-consistent generation. Compared to adapter-based approaches, this method deeply learns image-text alignment, fully leveraging foundation models and resolving degradation issues from multiple adapters.

\subsection{Subject-consistency video generation}
\label{sec:related_work:video}

From recent research developments, the advancement of video generation capabilities and algorithmic innovations tends to lag behind image tasks. Similar to image consistency techniques, Kling \cite{Keling} has released an optimization-based video generation method for facial ID consistency, which requires uploading multiple videos of the same person for optimization, resulting in significant computational costs. Adapter-based approaches have also been attempted for video ID consistency tasks, such as ID-Animator \cite{he2024id} and ConsisID \cite{yuan2024identity}. However, these works have been validated on small datasets (around 10k), which limits their ability to fully align facial information with text descriptions. Recent works like ConceptMaster \cite{huang2025conceptmaster}, MovieWeaver \cite{liang2025movie}, and VideoAlchemist \cite{chen2025multi} have demonstrated capabilities in generating consistent multi-subject videos in general scenarios. However, there are currently no open-source methods for the S2V task, and commercial software's S2V capabilities \cite{Vidu, Pika, Keling, Hailuo, polyak2024movie} remain state-of-the-art. Therefore, comparing the performance with commercial closed-source solutions is crucial for evaluating the superiority of the proposed method.

\section{Phantom}
\label{sec:phantom}

This section introduces the specific implementation of \textbf{\textit{Phantom}}. The first subsection describes how to construct cross-modal alignment training data, emphasizing the creation of cross-pair text-image-video triplets to address the "copy-paste" issue. The second subsection presents the design and considerations of the \textit{Phantom} architecture, focusing on how single and multiple subject features are dynamically injected into the framework. The third subsection introduces some key training settings and inference techniques to ensure the efficient implementation of S2V capabilities.

\subsection{Data Pipeline}
\label{sec:method:data_pipeline}

\begin{figure}[t]
	\centering
	\includegraphics[width=0.95\columnwidth]{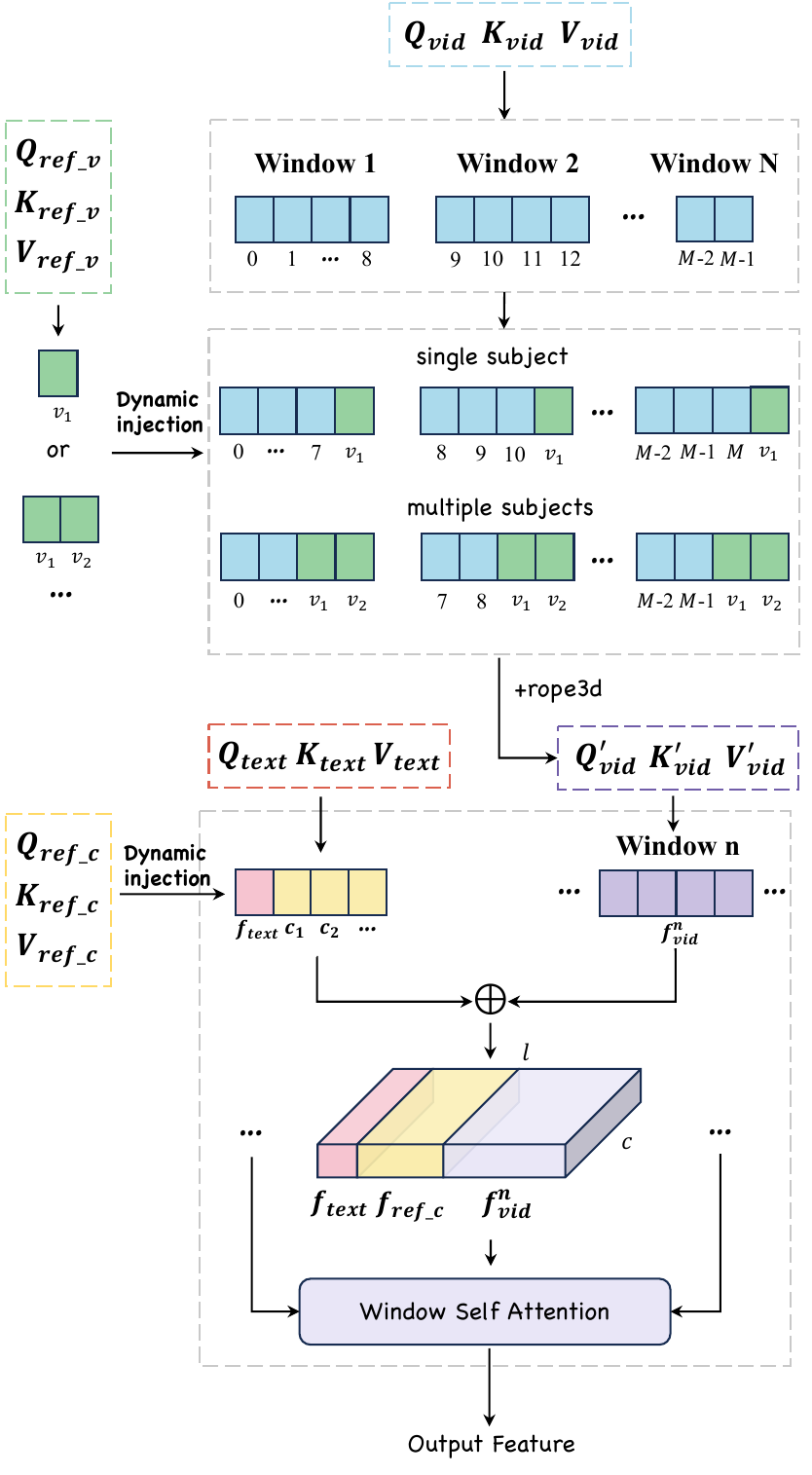} 
	\caption{\textbf{Dynamic injection strategy} and attention calculation for single or multiple reference subjects in each MMDiT block.}
	\label{fig:attention}
\end{figure}

To achieve subject-to-video (S2V) generation, we constructed a triplet data structure of text-image-video for cross-modal learning (Figure \ref{fig:data_pipeline}), ensuring that videos are paired with both images and text. 
First, we sampled long videos from Panda70M \cite{chen2024panda70m} and in-house sources. These videos were cut into single-scene segments using AutoShot \cite{zhu2023autoshot} and PySceneDetect \cite{pyscenedetect}, and any clips with low quality, aesthetics, or motion levels were filtered out.
Next, we used Gemini \cite{team2023gemini} to generate captions for the filtered video clips, focusing on describing the subjects' appearance, behavior, and the scene. 
Further, the LLM \cite{GPT4} is utilized to analyze the caption and extract the subject words with appearance descriptions, which are used as prompts for the VLM \cite{Qwen2.5-VL} to obtain the subject detection boxes of the reference frames.
At this point, the descriptions of the subjects in the captions can be exactly aligned with the subject elements detected in the reference images.

Although the reference images and text are aligned, the reference images are taken from specific frames within the videos. These image-video pairs are termed "in-pair" data. Some existing methods \cite{chen2025multi,huang2025conceptmaster} use in-pair data to train S2V models, ensuring subject consistency between images and videos. However, high visual similarity might cause the model to disregard text prompts, resulting in generated videos that simply copy-paste the input images. To address this issue, we undertake an additional effort to further establish pairings between cross-video clips. We employ the image embedder \cite{shao20221st} of the improved CLIP architecture to score and pair subjects detected across different videos. Pairs with scores that are excessively high (indicating a likelihood of copy-pasting) or too low (indicating different subjects) are eliminated.  

After constructing the cross-paired data pipeline, further segmentation is required based on application scenarios. 
These primary elements include people, animals, objects, backgrounds, and more. Additionally, interactions between multiple elements can further categorize scenarios, such as multi-person interactions, human-pet interactions, and human-object interactions. 
By segmenting the data sources according to these application scenarios, we can quantitatively supplement missing data types. For example, virtual try-on applications require specific collections of model images and garment layouts. 
Ultimately, we obtained cross-pair data on the order of one million, among which the data containing human subjects accounted for the largest proportion. In addition, we also added a portion of paired image data to increase diversity. The data sources are Subject200k \cite{chen2024unireal} and OmniGen \cite{xiao2024omnigen}.

\subsection{Framework}
\label{sec:method:framework}

\begin{figure*}[t]
	\centering
	\includegraphics[width=0.85\textwidth]{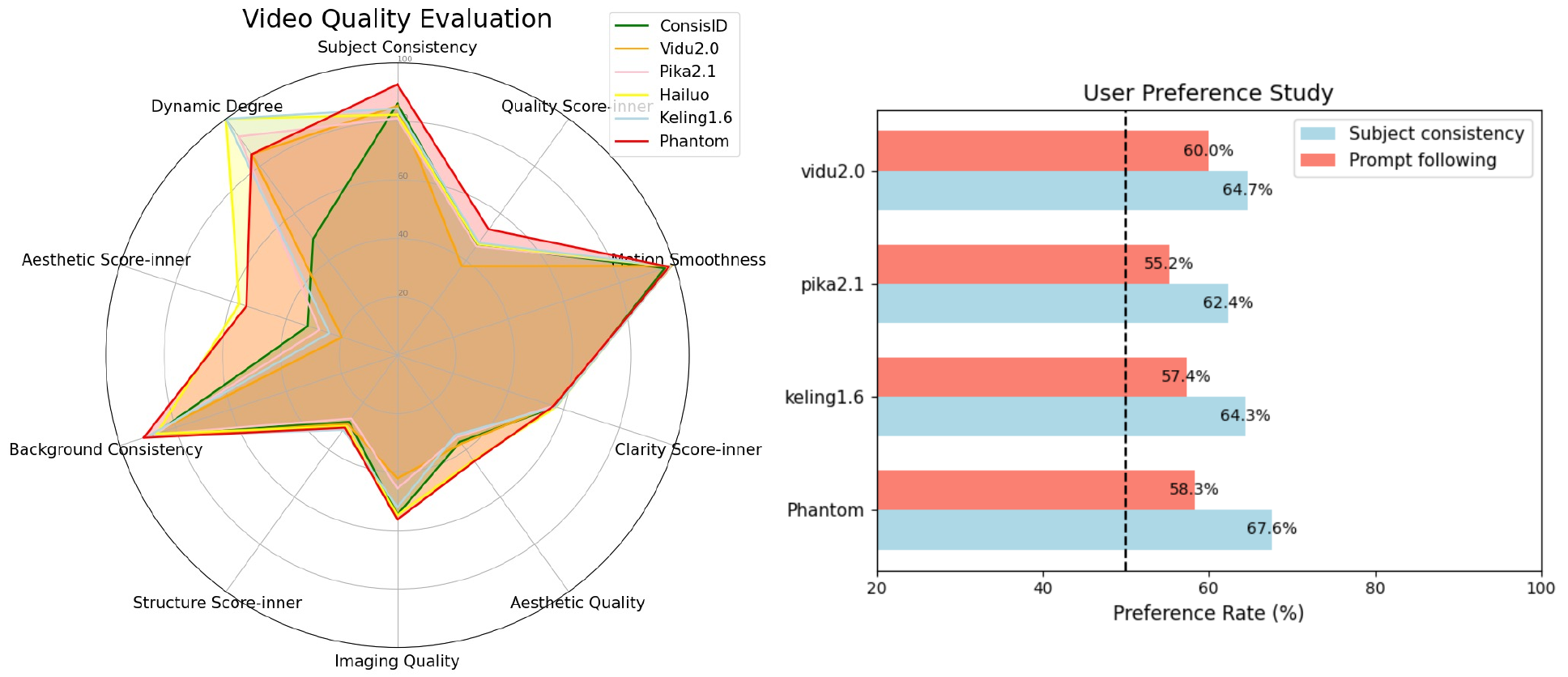} 
	\caption{Video quality evaluation (left) and user study results for multi-subject consistency (right).}
	\label{fig:radar}
\end{figure*}

The \textbf{\textit{Phantom}} architecture, shown in Figure \ref{fig:framework}, consists of an untrained input head and a trained MMDiT module. The input head includes a 3D VAE \cite{yang2024cogvideox}  encoder and an LLM \cite{yang2024qwen2} inherited from the video foundation model \cite{lin2025diffusion, wang2025seedvr}, which encode the input video and text, respectively. 
The vision encoder, critically, comprises both a Variational Autoencoder (VAE) \cite{esser2021taming} and CLIP \cite{zhai2023sigmoid, radford2021learning}. The image features $F_{\text{ref\_v}}$ concatenated with the video latents $F_{\text{vid}}$ reuse the 3D VAE to maintain consistency in the visual branch input. Meanwhile, the image CLIP features $F_{\text{ref\_c}}$ concatenated with the text features $F_{\text{text}}$ provide high-level semantic information, compensating for the low-level features from the VAE. Feature merging involves dimensional alignment, as detailed below, 
\begin{equation}
	F_{T}^{l_1 + l_2, c} =F_{\text{text}}^{l_1,c}\oplus F_{\text{ref\_c}}^{l_2,c},
\end{equation}
\begin{equation}
	F_{V}^{t+n,h,w,c} = F_{\text{vid}}^{t,h,w,c}\oplus F_{\text{ref\_v}}^{n,h,w,c},
	\label{eq:fV}
\end{equation}
where $\oplus$ denotes concatenation. The concatenated features $F_T$ and $F_V$ are fed into the visual and text branches of MMDiT, and the model only separates the injected features during the calculation of attention.

Specifically, the MMDiT block is based on \cite{lin2025diffusion, wang2025seedvr} and improved for reference image input, primarily modifying the Attention \cite{vaswani2017attention} block, as shown in Figure \ref{fig:attention}. 
First, the $Q_{\text{vid}}$, $K_{\text{vid}}$, $V_{\text{vid}}$ features calculated from $F_{\text{vid}}$ are divided into windows of size 9. 
Then, the $Q_{\text{ref\_v}}$, $K_{\text{ref\_v}}$, $V_{\text{ref\_v}}$ features calculated from $F_{\text{ref\_v}}$
are dynamically concatenated to the end of each window, while the in-situ features are sequentially shifted to the start of the next window. 
This approach maintains the window structure while ensuring interaction between video and subject features within each window, as well as adaptive input for single- or multi-subject. 
Meanwhile, the $Q_{\text{text}}$, $K_{\text{text}}$, $V_{\text{text}}$ features calculated from $F_{\text{text}}$ and the $Q_{\text{ref\_c}}$, $K_{\text{ref\_c}}$, $V_{\text{ref\_c}}$ features calculated from $F_{\text{ref\_c}}$ are dynamically concatenated. After collecting all reference information, self-attention is calculated within each window. Then, the dynamically injected reference image features (including $\text{ref\_v}$ and $\text{ref\_c}$) and the text features within each window are extracted from the output features and averaged. This process ensures that the dimensions of the input and output features within the current block remain consistent, thereby facilitating subsequent block computations.

\begin{figure*}[t]
	\centering
	\includegraphics[width=\textwidth]{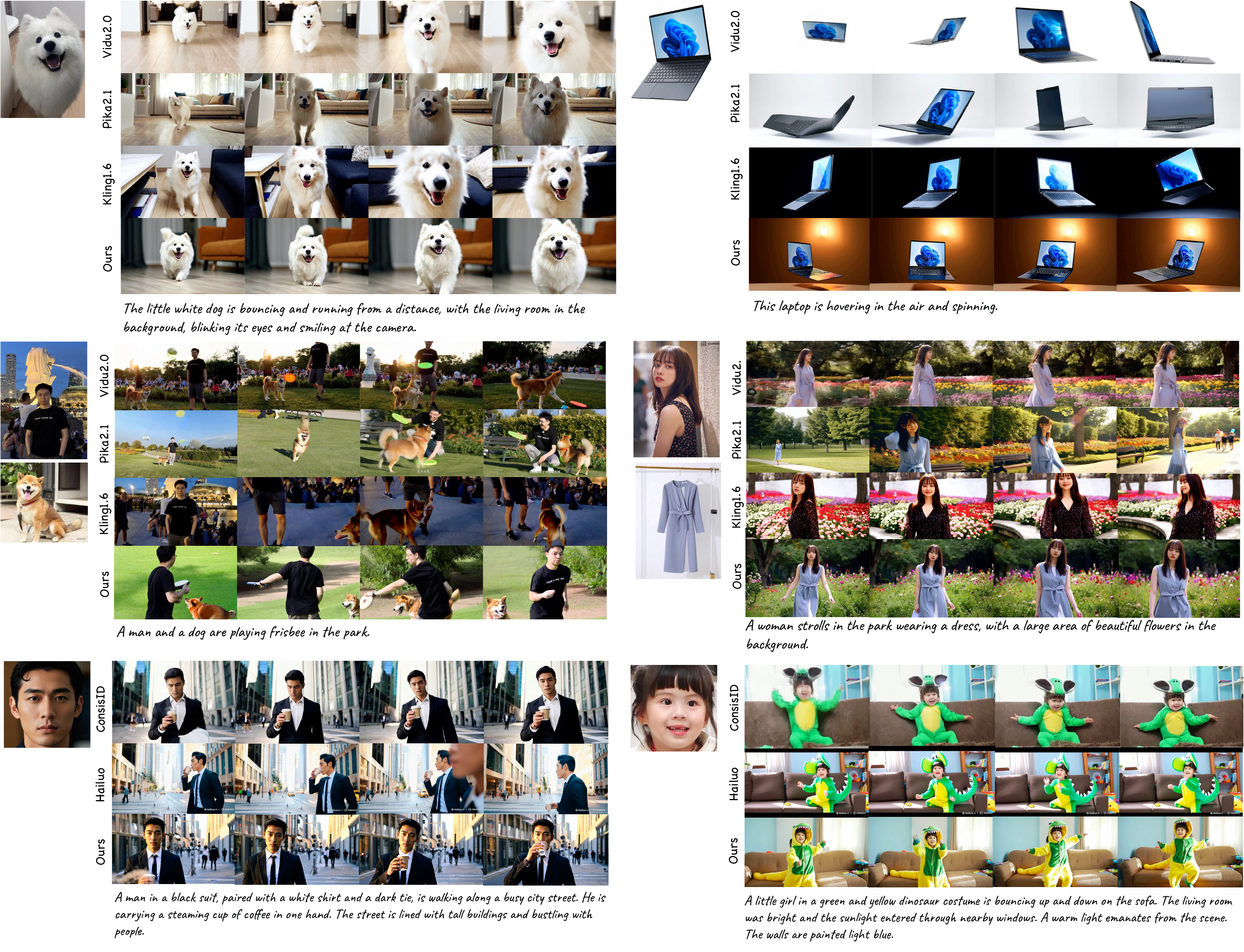} 
	\caption{Comparison results showing, from top to bottom, single subject, multi-subject, and facial ID-consistent video generation, with four uniformly sampled frames displayed in each case.}
	\label{fig:ipid_eval}
\end{figure*}

\subsection{Training and inference}
\label{sec:method:train}

\textbf{Training setup.} 
We employ rectified flow (RF) \cite{liu2022flow,lipman2022flow} to construct the training objective and adjust the noise distribution sampling \cite{esser2024scaling}. RF aims to learn an appropriate flow field, enabling the model to efficiently and high-quality generate meaningful data samples from noise.
In the forward process of training, noise is added to clean data $x_0$ to generate $x_t = (1-t) \cdot x_0 + t \cdot \epsilon $, where $\epsilon$ is Gaussian noise with the distribution $\mathcal{N}(0, \text{I})$ and $t$ is a randomly sampled step scaled to a value between 0 and 1 based on the total steps (T=1000). 
The model predicts velocity $v_t$ to regress velocity $u_t = dx_t/dt$, and $v_t$ is represented by,
\begin{equation} 
v_t = \mathcal{G}_\theta (x_t, t, F_T, F_V).
\end{equation}
Thus, the RF training loss is given by,
\begin{equation} 
\mathcal{L}_{\text{mse}} = {\left \| v_t  - u_t \right \|}^{2}.
\end{equation} 
Notably, $v_t$ includes additional (n)-dimensional features at the tail (refer to Eq.\ref{eq:fV}), which does not participate in the loss calculation.
The model training is conducted in two phases: the first phase trains for 50k iterations at 256p/480p resolution, and the second phase incorporates mixed 720p data, training for an additional 20k iterations to enhance higher resolution generation capabilities. Additionally, since one of the training objectives of VAE is pixel-level reconstruction, the CLIP features can be overshadowed when trained together with VAE features. Therefore, we set a relatively high dropout rate (0.7) for VAE during training to achieve balance. The total computational resources consumed approximately 30,000 GPU-hours on A100.

\noindent \textbf{Inference settings.}
\textit{Phantom} inference can accept 1 to 4 reference images and generate corresponding videos by describing the reference subjects using a given text prompt. Note that generating with more reference subjects may lead to unstable results. 
To align with the training data, the text prompt used in inference must first be adjusted by a rephraser to ensure it accurately describes the appearance and behavior of each reference subject, avoiding confusion between similar subjects (see supplementary materials). 
The Euler method is used for sampling over 50 steps, and the classifier-free guidance \cite{ho2022classifier} separates the image and text conditions. The denoised output at each step is given by,
\begin{equation}
	\begin{split}
		x_{t-1} = & x_{t-1}^{\oslash} + \omega_1 ( x_{t-1}^I - x_{t-1}^{\oslash} ) \\
		& + \omega_2 ( x_{t-1}^{TI} - x_{t-1}^{I} ),
	\end{split}
	\label{eq:inference}
\end{equation}
where $x_{t-1}^{\oslash}$ is the unconditional denoising output, $x_{t-1}^I$ is the image-conditioned denoising output, and $x_{t-1}^{TI}$ is the joint text-image conditioned denoising output. The weights $\omega _1$ and $\omega _2$ are set to 3 and 7.5, respectively.

\section{Experiments}
\label{sec:exps}

\subsection{Evaluation materials}
\label{sec:exp:eval_mat}

\textit{Phantom} can be fine-tuned from any video generation base model \cite{lin2025diffusion, wang2025seedvr}. The T2V and I2V pre-training stages are excluded from this evaluation. We focus on assessing the subject consistency generation capability, with additional independent evaluations for face ID-based video generation. Due to the lack of an established benchmark for subject-to-video, we constructed a specific test set and defined evaluation metrics accordingly.

We collected 50 reference images from different scenarios, covering humans, animals, products, environments, and clothing. Each reference image is paired with 3 different text prompts. To ensure confidence in each case, each text-image pair is generated with three random seeds, resulting in a total of 450 videos. For scenarios with multiple reference images, we mixed the aforementioned reference images and rewrote the text prompts to obtain a test set of 50 groups. Additionally, considering the unique value of portrait scenarios, we collected an additional 50 portrait reference images, including both celebrities and ordinary individuals, for independent evaluation of ID consistency. 

For the S2V task, the existing available state-of-the-art (SOTA) methods are closed-source commercial tools. Therefore, we evaluated and compared the latest capabilities of Vidu \cite{Vidu}, Pika \cite{Pika}, and Kling \cite{Keling}. For the ID-preserving video generation task, the commercial tool Hailuo \cite{Hailuo} demonstrated impressive results. We also evaluated an excellent open-source algorithm ConsisID \cite{yuan2024identity}.

\subsection{Quantitative results}
\label{sec:exp:quant}

We classify the S2V evaluation metrics into three major categories: video quality, text-video consistency, and subject-video consistency. First, the visualization of video quality is shown in the radar chart on the left side of Figure \ref{fig:radar}. We selected six metrics provided by VBench \cite{huang2024vbench++}  for testing and supplemented them with four inner model scores such as structure breakdown score. For text-video consistency, we used ViCLIP \cite{wang2022internvideo} to directly calculate the cosine similarity score between the text and the video. For single subject consistency, we uniformly sampled 10 frames from each video and calculated the CLIP \cite{cherti2023reproducible} and DINO \cite{oquab2023dinov2} feature Direction Scores with the reference image. Additionally, we used grounded-sam to segment the subject part of the video and calculate the CLIP and DINO scores (excluding scene graphs). For ID consistency, we used three facial recognition models to measure similarity \cite{deng2019arcface, huang2020curricularface}. 

The video quality evaluation results, shown on the left side of Figure \ref{fig:radar}, indicate that \textit{Phantom} performs slightly worse \cite{huang2024vbench++}, while excelling in other metrics. As shown in Table \ref{tab:id_eval} and \ref{tab:ip_eval}, \textit{Phantom}  leads in overall metrics for subject consistency (Identity Consistency) and prompt following. For multi-subject video generation, due to high error rates in automated subject detection and matching, we conducted a user study. We surveyed 20 users, who rated the methods on a scale of 1 to 3 (1: unusable, 2: usable, 3: satisfactory). The evaluation results, displayed in the bar chart on the right side of Figure \ref{fig:radar}, show that \textit{Phantom} 's multi-subject performance is comparable to commercial solutions, with some advantages in subject consistency.

\begin{table}[t]
	\centering
	\resizebox{\columnwidth}{!}{
		\small
		\begin{tabular}{lcccc}
			\toprule
			\multirow{2}{*}{Methods} & \multicolumn{3}{c}{Identity Consistency} & \multicolumn{1}{c}{Prompt Following} \\
			\cmidrule(lr){2-4} \cmidrule(lr){5-5}
			& FaceSim-Arc $\uparrow$ & FaceSim-Cur $\uparrow$ & FaceSim-glink $\uparrow$ & ViCLIP-T $\uparrow$ \\
			\midrule
			ConsisID                  & 0.538       & 0.417       & 0.470         & 21.76    \\
			Hailuo-ID                 & 0.542       & 0.504       & 0.557         & 23.31    \\
			Phantom-ID                & \textbf{0.581} & \textbf{0.529} & \textbf{0.590} & \textbf{24.12}   \\
			\bottomrule
		\end{tabular}
	}
	\caption{Comparison of different methods based on identity consistency and prompt following}
	\label{tab:id_eval}
\end{table}

\begin{table}[t]
	\centering
	\resizebox{\columnwidth}{!}{
		\small
		\begin{tabular}{lcccccc}
			\toprule
			\multirow{2}{*}{Methods} & \multicolumn{4}{c}{Subject Consistency} & \multicolumn{1}{c}{Prompt Following} \\
			\cmidrule(lr){2-5} \cmidrule(lr){6-6}
			& CLIP-I $\uparrow$ & DINO-I $\uparrow$ & CLIP-I-Seg $\uparrow$ & DINO-I-Seg $\uparrow$ & ViCLIP-T $\uparrow$ \\
			\midrule
			Vidu2.0            & 0.706 & 0.511 & \underline{0.724} & \textbf{0.544}  & 22.78 \\
			Pika2.1            & 0.697 & 0.498 & 0.712 & 0.534  & \underline{23.05} \\
			Kling1.6          & \textbf{0.732} & \textbf{0.554} & 0.715 & 0.569 & 21.62 \\
			Phantom-IP         & \underline{0.714} & \underline{0.523} & \textbf{0.731} & \underline{0.538} & \textbf{23.41} \\
			\bottomrule
		\end{tabular}
	}
	\caption{Comparison of different methods based on single subject consistency and prompt following. \textbf{Boldface} indicates the highest scores in each column, and \underline{underline} indicates the second-highest scores.}
	\label{tab:ip_eval}
\end{table}

\subsection{Qualitative results}
\label{sec:exp:qual}

We present the comparison results of several typical cases in Figure \ref{fig:ipid_eval}. Each generated video is displayed with four evenly sampled frames, including the first and last frames. The first two rows of Figure \ref{fig:ipid_eval}  respectively show the results of generating single- and multiple subject consistency. It can be seen that Vidu \cite{Vidu} and \textit{Phantom} exhibit balanced performance in subject consistency, visual effect, and text response. Pika \cite{Pika} performs poorly in subject consistency. Kling \cite{Keling} has a notable issue: some cases exhibit characteristics analogous to I2V approaches. For instance, the first frame of character videos almost matches the input reference image, leading to low success rates in virtual try-on scenarios. Additionally, the laptop case shows that the compared methods tend to cause deformations in rigid body movements. The last row of Figure \ref{fig:ipid_eval} shows the results of video generation for facial ID preservation. The open-source method ConsisID \cite{yuan2024identity} tends to exhibit motion blur, and has weak text response. Hailuo \cite{Hailuo} excels in visual aesthetics, but there is some loss in facial similarity. Our results are balanced across all dimensions, with particular advantage in ID consistency. More qualitative analyses are presented in supplementary materials.

\subsection{Ablation study}
\label{sec:exp:ablation}

\begin{table}[t]
	\centering
	\resizebox{\columnwidth}{!}{
	\small
	\begin{tabular}{lcccccc}
		\toprule
		\multirow{2}{*}{\vtop{\hbox{\strut Methods}}} & \multicolumn{2}{c}{Subject Consistency} & \multicolumn{1}{c}{Prompt Following} & \multicolumn{2}{c}{Video Quality} \\
		\cmidrule(lr){2-3} \cmidrule(lr){4-4} \cmidrule(lr){5-6}
		& CLIP-I $\uparrow$ & DINO-I $\uparrow$ & ViCLIP-T $\uparrow$ & Aes score $\uparrow$ & Clarity score $\uparrow$ \\
		\midrule
		w/o CLIP & 0.693 & 0.519 & \textbf{23.63} & 62.03 & 71.40 \\
		w/o VAE & 0.512 & 0.302 & 22.79 & 48.82 & 70.76 \\
		w/ All & \textbf{0.714} & \textbf{0.523} & \underline{23.40} & \textbf{64.32} & \textbf{71.72} \\
		\bottomrule
	\end{tabular}}
	\caption{The ablation experiment results of VAE and CLIP.}
	\label{tab:vae_clip}
\end{table}

\begin{figure}[t]
	\centering
	\includegraphics[width=0.98\columnwidth]{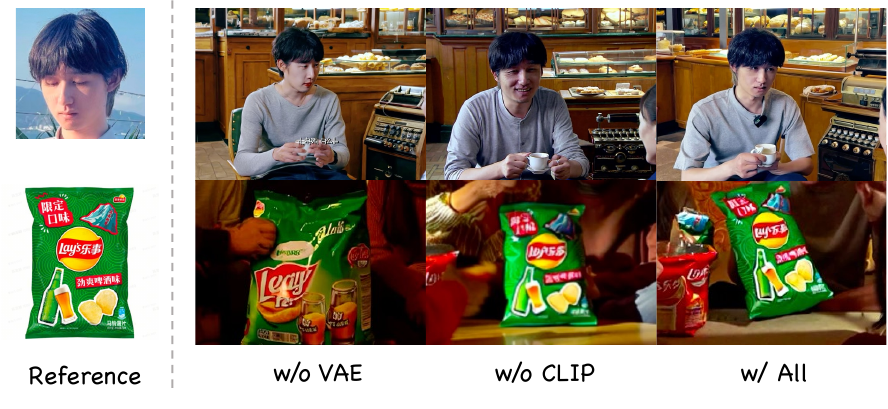} 
	\caption{Qualitatively display the ablation of VAE and CLIP.}
	\label{fig:vae_clip}
\end{figure}

\noindent \textbf{Selection of visual encoder.} 
Due to differences in training methods, CLIP aligns image-text pairs and tends to extract semantic information, while VAE aims for lossless reconstruction and focuses on detailed information. As shown in Figure \ref{fig:vae_clip}, faces generated using only CLIP features are smoother and more refined but show decreased similarity. In contrast, faces generated using VAE features are sharper but may amplify undesirable details, making them contain more artifacts.
In general object scenarios, CLIP is inadequate at reproducing details like text and patterns, thus primarily serving to supplement VAE's high-level information. Quantitative results in Table \ref{tab:vae_clip} show that combining VAE and CLIP features is more advantageous. Additional ablation studies are given in supplementary materials.

\section{Conclusion}
\label{sec:concluson}

We propose \textbf{\textit{Phantom}}, a method for subject-consistent video generation that achieves cross-modal alignment through text-image-video triplet learning. By redesigning the joint text-image injection mechanism and leveraging dynamic feature integration, \textit{Phantom} demonstrates competitive performance in unified single/multi-subject generation and facial ID preservation tasks, outperforming commercial solutions in quantitative evaluations.

\noindent \textbf{Acknowledgments}
We would like to express our gratitude to the Bytedance-Seed team for their support. Special thanks to Haoyuan Guo, Zhibei Ma, Sen Wang and Lu Jiang for their assistance with the model and data. In addition, we are also very grateful to Siying Chen, Qingyang Li, and Wei Han for their help with the evaluation.

{
    \small
    \bibliographystyle{ieeenat_fullname}
    \bibliography{main}
}
\clearpage
\appendix
\section*{Supplementary Materials}
\label{sec:supp}

\subsection*{1. S2I+I2V \textbf{\textit{vs}} S2V}
\label{sec:supp:s2i2v}
\begin{figure}[h]
	\centering
	\includegraphics[width=0.98\columnwidth]{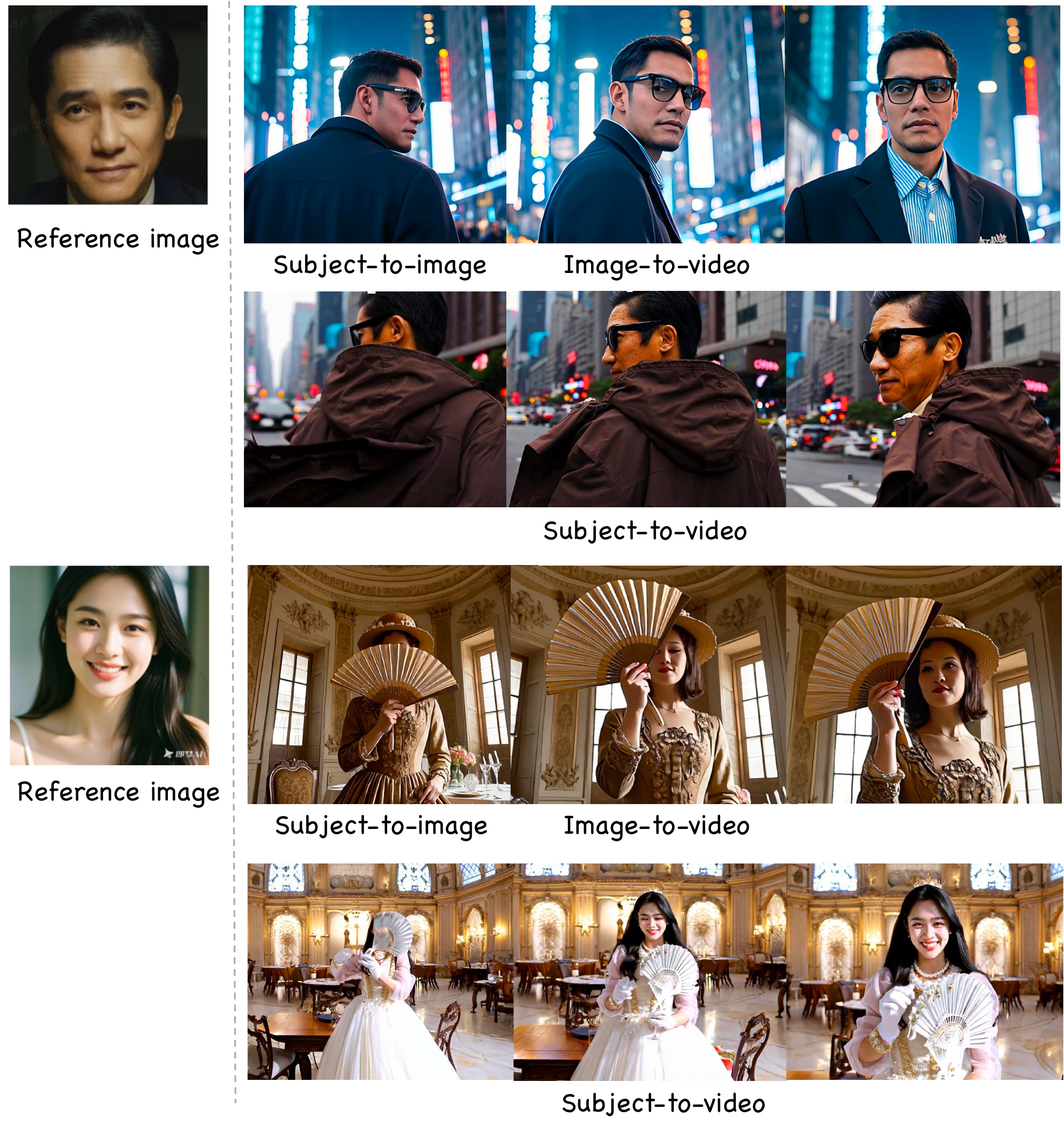} 
	\caption{Comparison of subject-to-image-to-video \cite{dreamina} and subject-to-video (ours).}
	\label{fig:s2i2v}
\end{figure}

As mentioned in the main text, combining subject-to-image (S2I) and image-to-video (I2V) can achieve similar effects to subject-to-video (S2V), but there are some difficult limitations. Firstly, existing methods \cite{guo2024pulid, huang2024realcustom, dreamina} for generating subject-consistent images or ID-consistent images still exhibit noticeable artificial artifacts, and there is significant room for improvement in the dimension of subject consistency. Equally important, I2V cannot ensure consistency of the subject during motion. As illustrated in Figure \ref{fig:s2i2v}, when inputting a reference portrait, S2I first generates a reference image for the initial frame of I2V. If the initial frame includes a back view or occlusions, I2V may "imagine" a false ID during the process of removing the occlusion, leading to a failure in maintaining consistency.

\subsection*{2. Copy-paste problem}
\label{sec:supp:copypaste}

\begin{figure}[h]
	\centering
	\includegraphics[width=0.98\columnwidth]{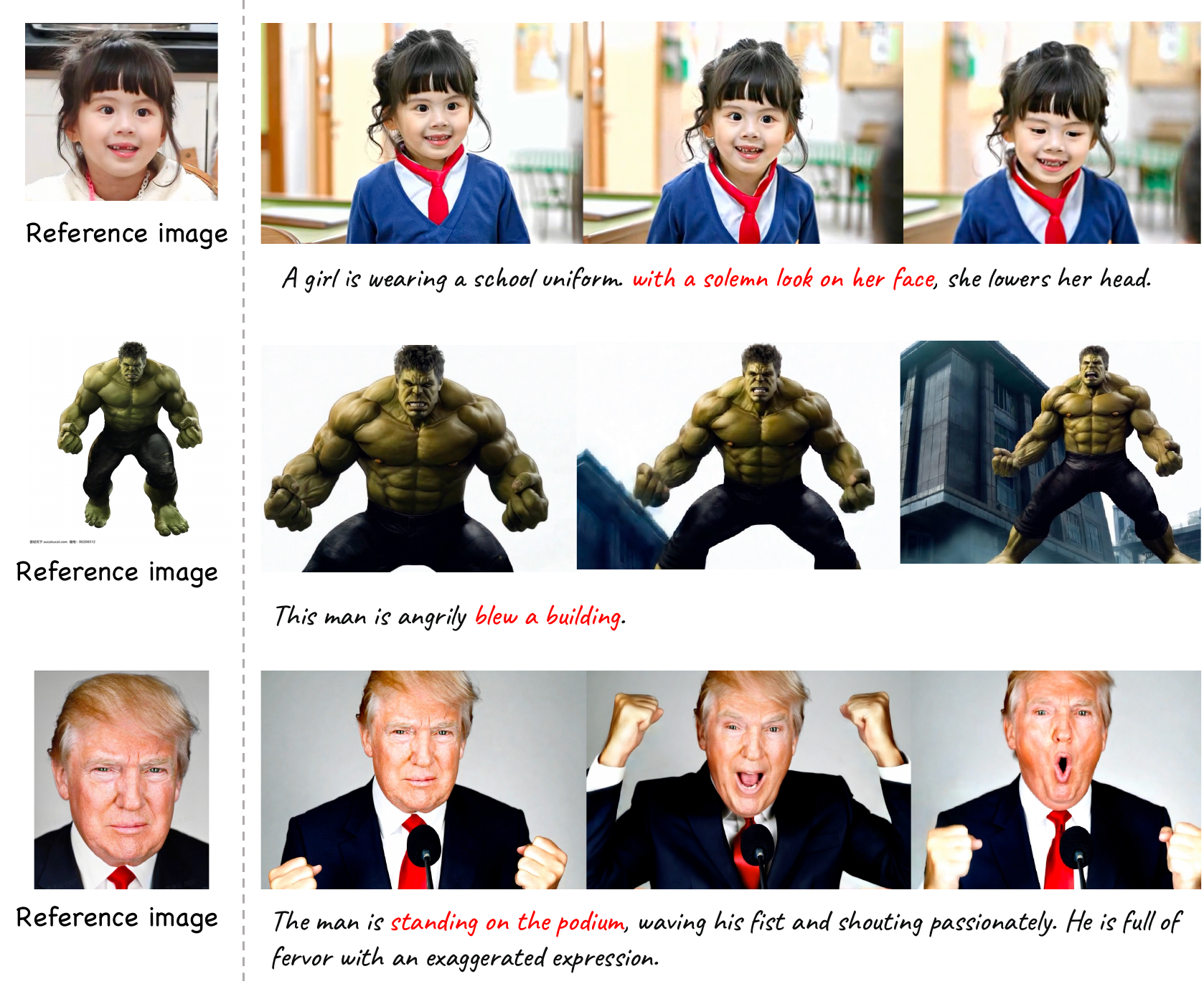} 
	\caption{Intuitive cases of copy-paste problems. The red font in the text prompt does not function as intended.}
	\label{fig:copypaste}
\end{figure}

In the field of video generation, the copy-paste issue is particularly prominent, manifesting as the leakage of image content into the generated video. Some methods sample keyframes from a video and use them as image conditions to reconstruct the video. However, this approach allows the model to employ shortcut learning strategies, simplifying the content understanding process. Figure \ref{fig:copypaste} shows examples of the copy-paste issue, sampling from the initial, middle, and final frames: In the first row, the girl's expression remains unchanged, ignoring the text prompt. In the second row, the cartoon character's movements remain stiff and identical to the reference. The third row illustrates a common case where the generated video is too similar to I2V, diminishing the effectiveness of scene-related text and reducing content diversity. To address this, we focus on constructing cross-video multi-subject pairings, ensuring subjects match in content while allowing for non-rigid deformations and changes in color distribution, thereby avoiding the copy-paste problem.

\begin{figure*}[t]
	\centering
	\includegraphics[width=\textwidth]{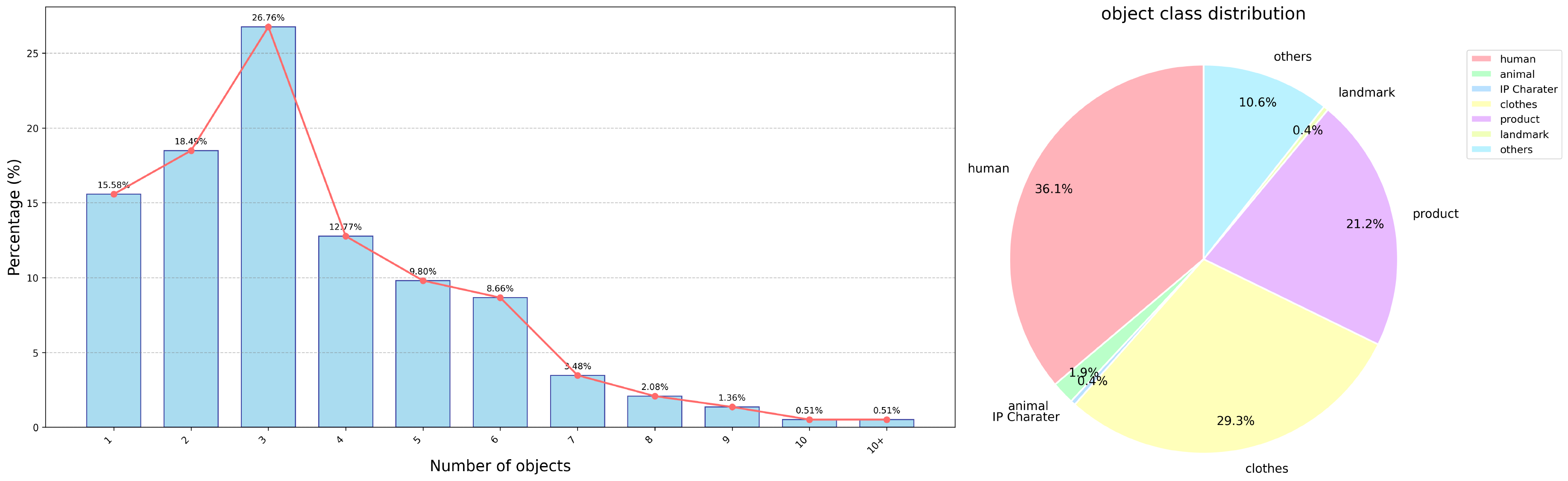} 
	\caption{Distribution of object frequencies and class.}
	\label{fig:data_dist}
\end{figure*}

\subsection*{3. Ablation study supplement}
\begin{figure}[t]
	\centering
	\includegraphics[width=0.98\columnwidth]{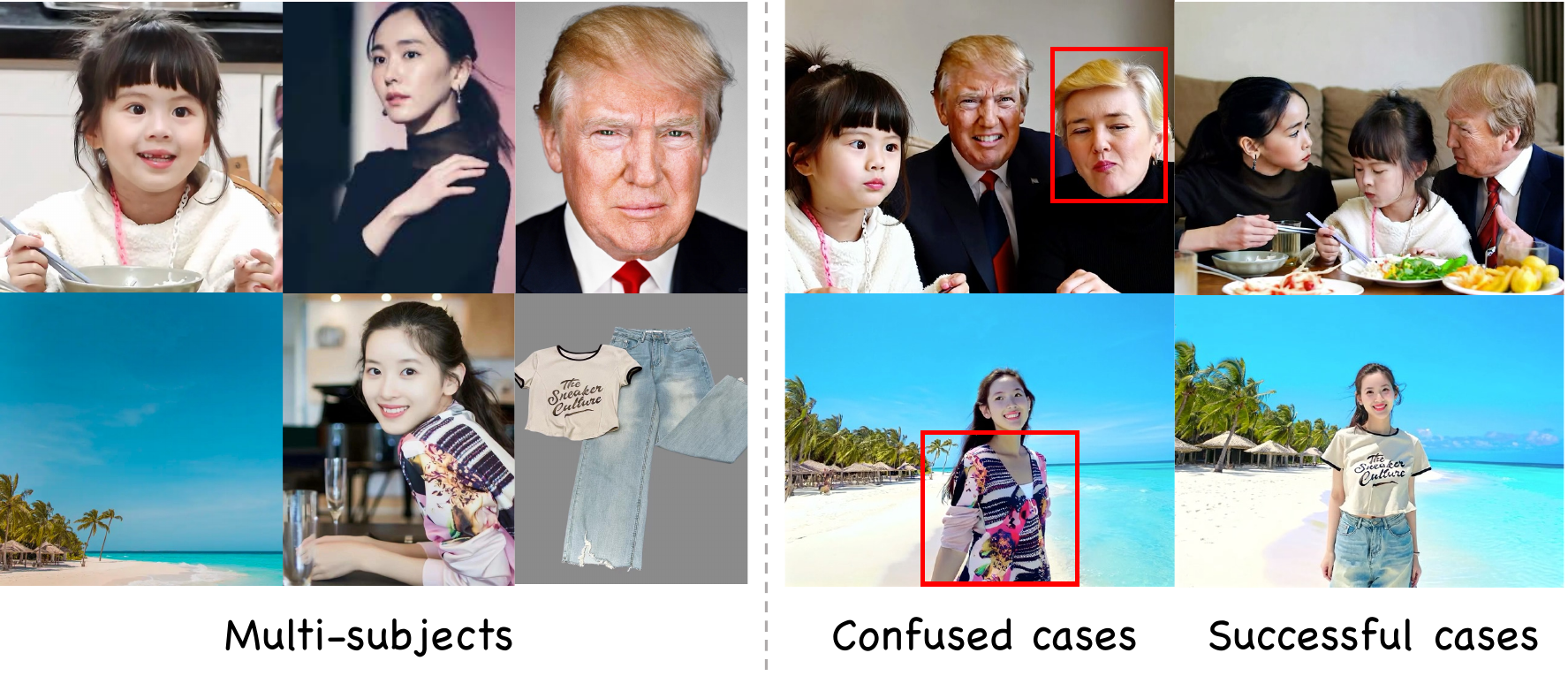} 
	\caption{Examples of multi-subject confusion: On the left are the multi-subject reference images, while the right columns present the cases of confusion and the successful cases after improvement.}
	\label{fig:confused}
\end{figure}

\begin{table}[b]
	\centering
	\resizebox{\columnwidth}{!}{
	\begin{tabular}{lcc}
		\toprule
		& \textbf{\textit{w/o}} text-image alignment & \textbf{\textit{w/}} text-image alignment \\
		\midrule
		Success rate & 65\% & 95\% \\
		\bottomrule
	\end{tabular}
	}
	\caption{Success rate of multi-subject generation with and without text-image alignment.}
	\label{tab:success_rate}
\end{table}

\textbf{Multi-subject confusion issue.} When multiple reference subjects are input simultaneously, appearance confusion may occur. Our solution aligns text descriptions with video subjects during training, ensuring distinct descriptions for each subject. During inference, a rephraser adjusts the input text prompts to align with the training data format. For example, in the first row of Figure \ref{fig:confused}, the original prompt "A family of three is having a meal at the table" caused confusion. The rephrased prompt "a woman in black, a young girl in white, and an elderly man in a suit eating together at the table" resolved this issue. In the second row of Figure \ref{fig:confused}, the original prompt "a girl in casual clothes walking by the beach" failed to match the reference. The rephrased prompt "a girl in a white T-shirt and jeans walking by the beach" successfully matched the reference.
Quantitative analysis, shown in Table \ref{tab:success_rate}, indicates a significant increase in the success rate of subject-consistent generation with this method.
Aligning image and text is crucial for multi-subject generation tasks. This approach, which requires no additional complex data structures or model designs, significantly optimizes the multi-subject confusion problem.

\subsection*{4. Data pipeline for face ID }
\label{sec:supp:facedata}

\begin{figure}[h]
	\centering
	\includegraphics[width=0.98\columnwidth]{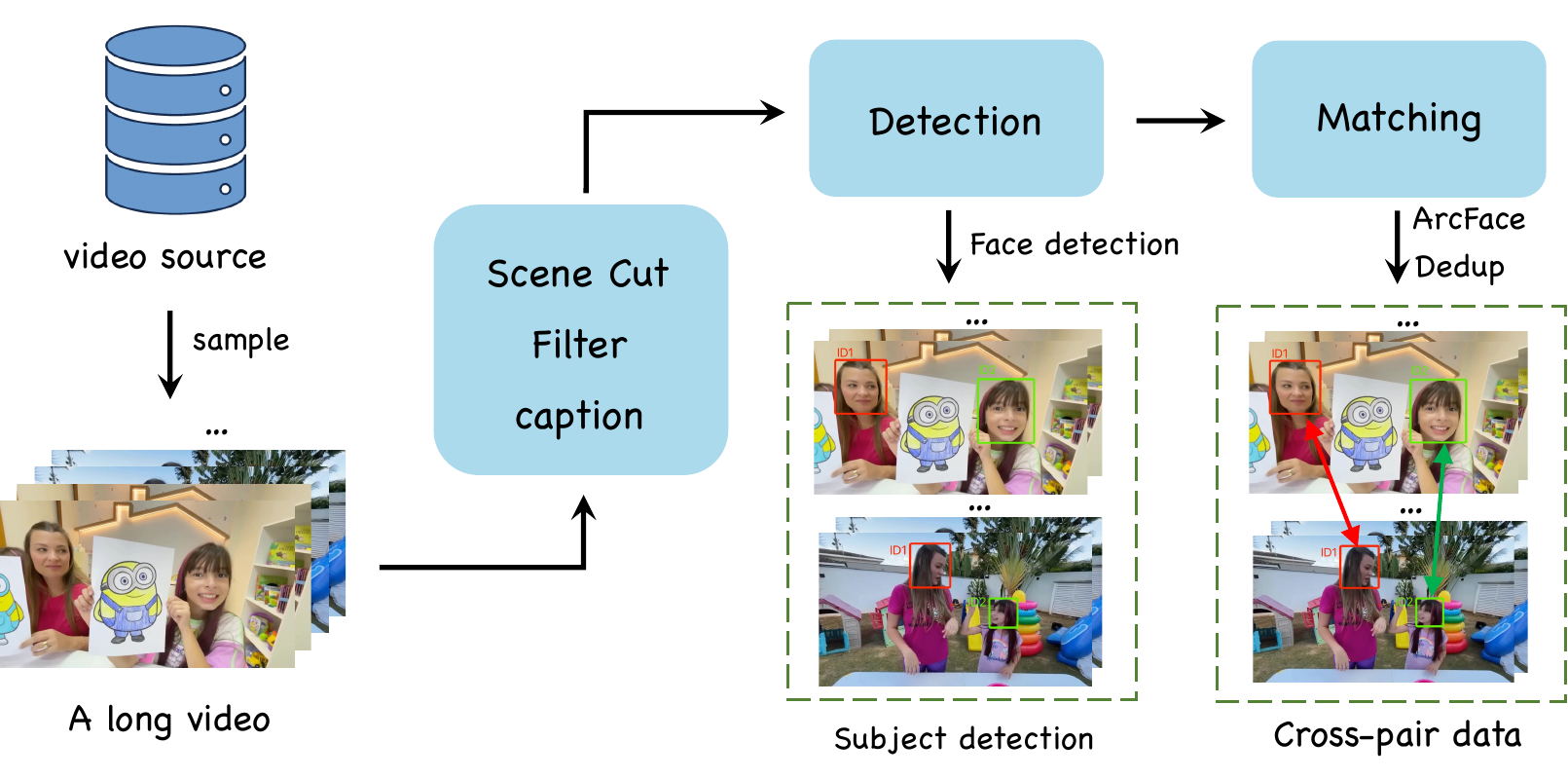} 
	\caption{Facial data processing pipeline for constructing ID cross-pair}
	\label{fig:facepipe}
\end{figure}

To enhance facial ID consistency, we developed an additional data pipeline for processing facial data. As shown in Figure \ref{fig:facepipe}, the facial data pipeline reuses the scene segmentation, video filtering, and annotation steps from the general subject pipeline. During the detection stage, we use an internal facial detection tool to identify each face in the video reference frames and calibrate it with the VLM \cite{Qwen2.5-VL} results from the captions using IOU (Intersection Over Union). In the matching stage, we calculate facial similarity using Arcface \cite{deng2019arcface} features and add a deduplication operator \cite{idealods2019imagededup} to further calibrate the recognition results.

\subsection*{5. Data distribution}
\textbf{Distribution of video object quantities.} We sample three frames at [0.05, 0.5, 0.95] of the video timeline and perform object detection on these frames. We filter out objects that meet the following criteria: (1) objects that are small in size or occupy a small proportion of the frame; (2) objects with a high degree of overlap with other objects; and (3) incomplete objects judged by the VLM \cite{Qwen2.5-VL}. The final distribution of the number of objects per video is shown in the table on the left side of Figure \ref{fig:data_dist}.

\noindent \textbf{Distribution of video object types.} We use LLM \cite{GPT4} to classify the noun fields in all captions into the following categories: human, animal, clothes, product, landmark, IP character, and others. The distribution is shown in the accompanying Figure \ref{fig:data_dist}, with human, clothes, and product categories accounting for the majority.

\subsection*{6. Model architecture}
\label{sec:supp:model}

\begin{figure}[t]
	\centering
	\includegraphics[width=\columnwidth]{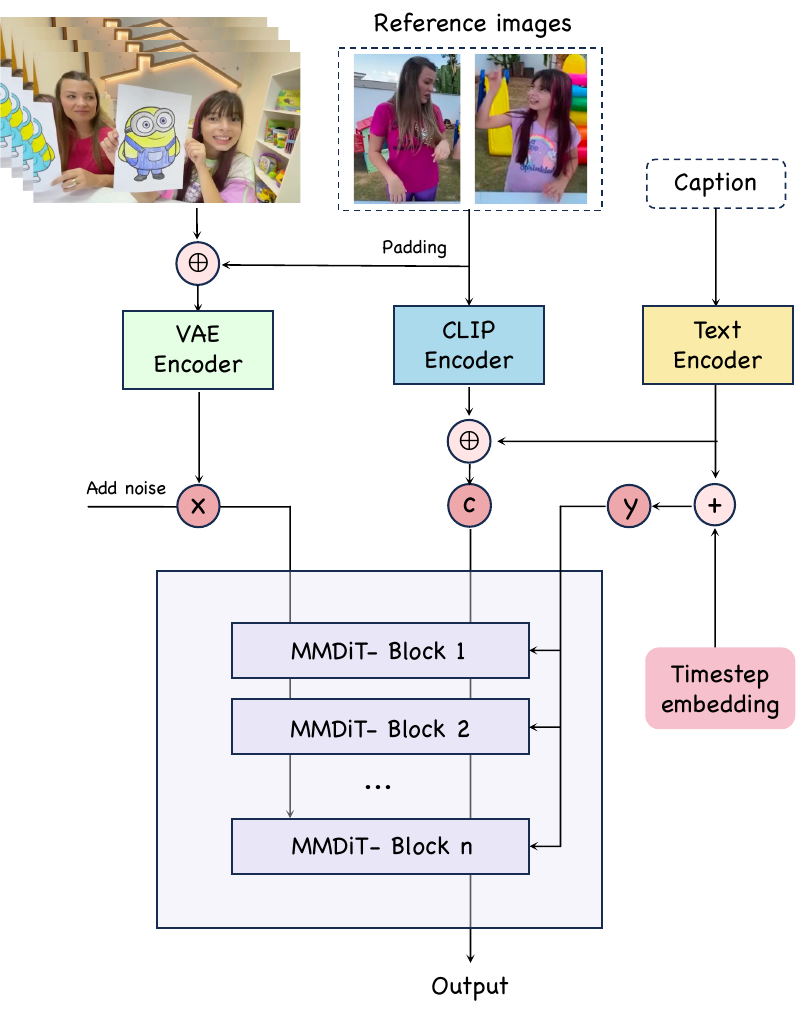} 
	\caption{The supplementary diagram of the Phantom framework.}
	\label{fig:architecture}
\end{figure}
The architecture of the \textit{Phantom} model is shown in Figure \ref{fig:architecture}, which supplements the missing details in the main text. As illustrated, it integrates the VAE and CLIP encoders to process reference images, while the text encoder handles captions. The encoded features are combined with added noise and processed through multiple MMDiT blocks, resulting in the final output. This design ensures a balance between detailed reconstruction and high-level information alignment while also guaranteeing a unified training paradigm for single and multiple subject inputs.

\subsection*{7. Qualitative analysis}
\label{sec:supp:qualitative}

\begin{figure*}[t]
	\centering
	\includegraphics[width=0.78\textwidth]{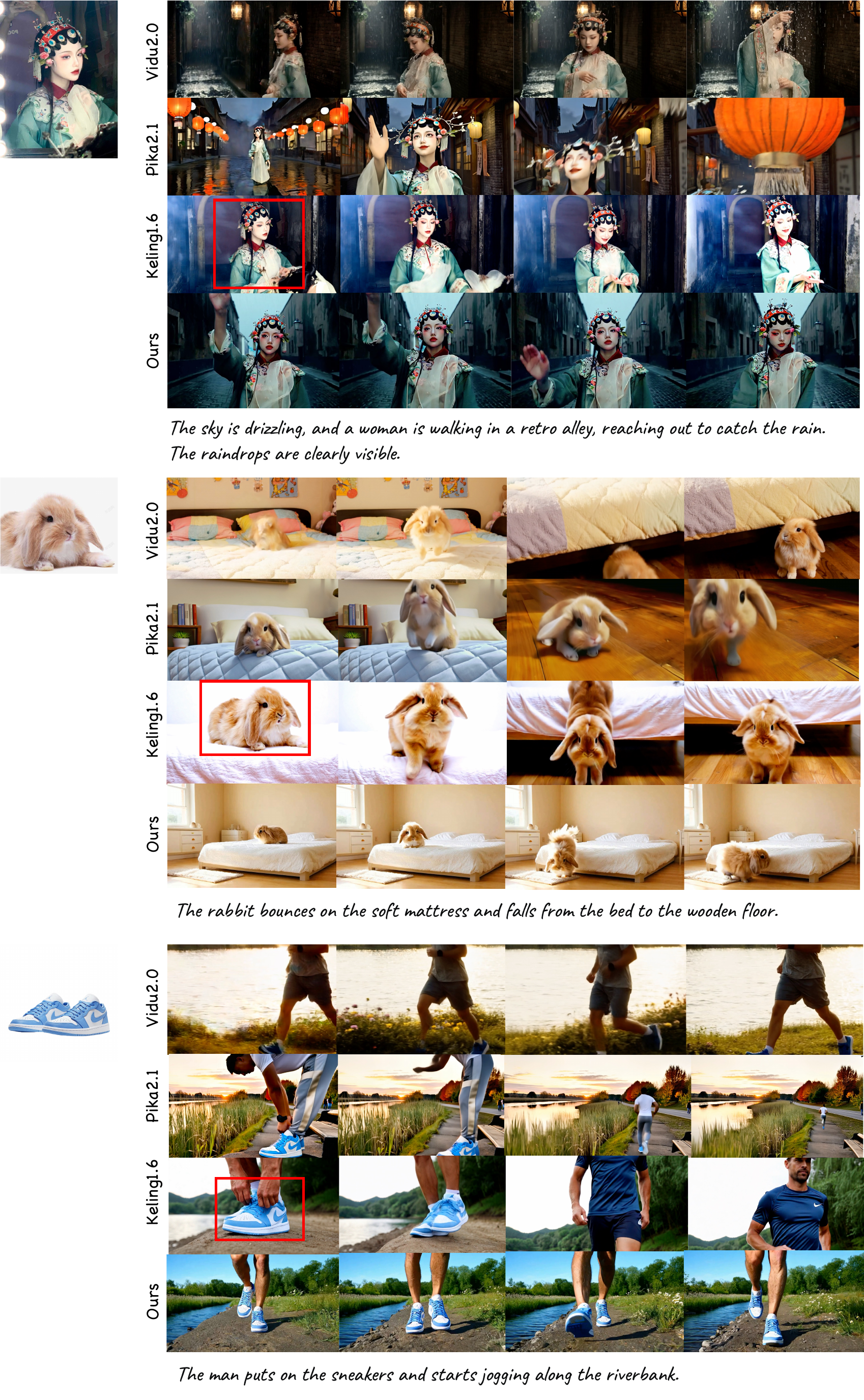} 
	\caption{Comparative results of single reference subject-to-video generation.}
	\label{fig:supp_sip}
\end{figure*}

Qualitative comparison results of single-subject consistency generation are shown in Figure \ref{fig:supp_sip}. Firstly, Vidu \cite{Vidu} performs well in both image consistency and text following for the first two cases but fails in the third shoe case with two different seeds. The effectiveness of Pika \cite{Pika} is evident, as the first two cases show significant disadvantages in maintaining subject consistency, tending towards a cartoonish appearance. The major issue with Kling \cite{Keling} is that most cases resemble the I2V mode, where the initial frame directly replicates the reference image (as indicated by the red box in Figure \ref{fig:supp_sip}), followed by subject motion generated based on text, thereby limiting the effectiveness of textual descriptions.

\begin{figure*}[t]
	\centering
	\includegraphics[width=0.78\textwidth]{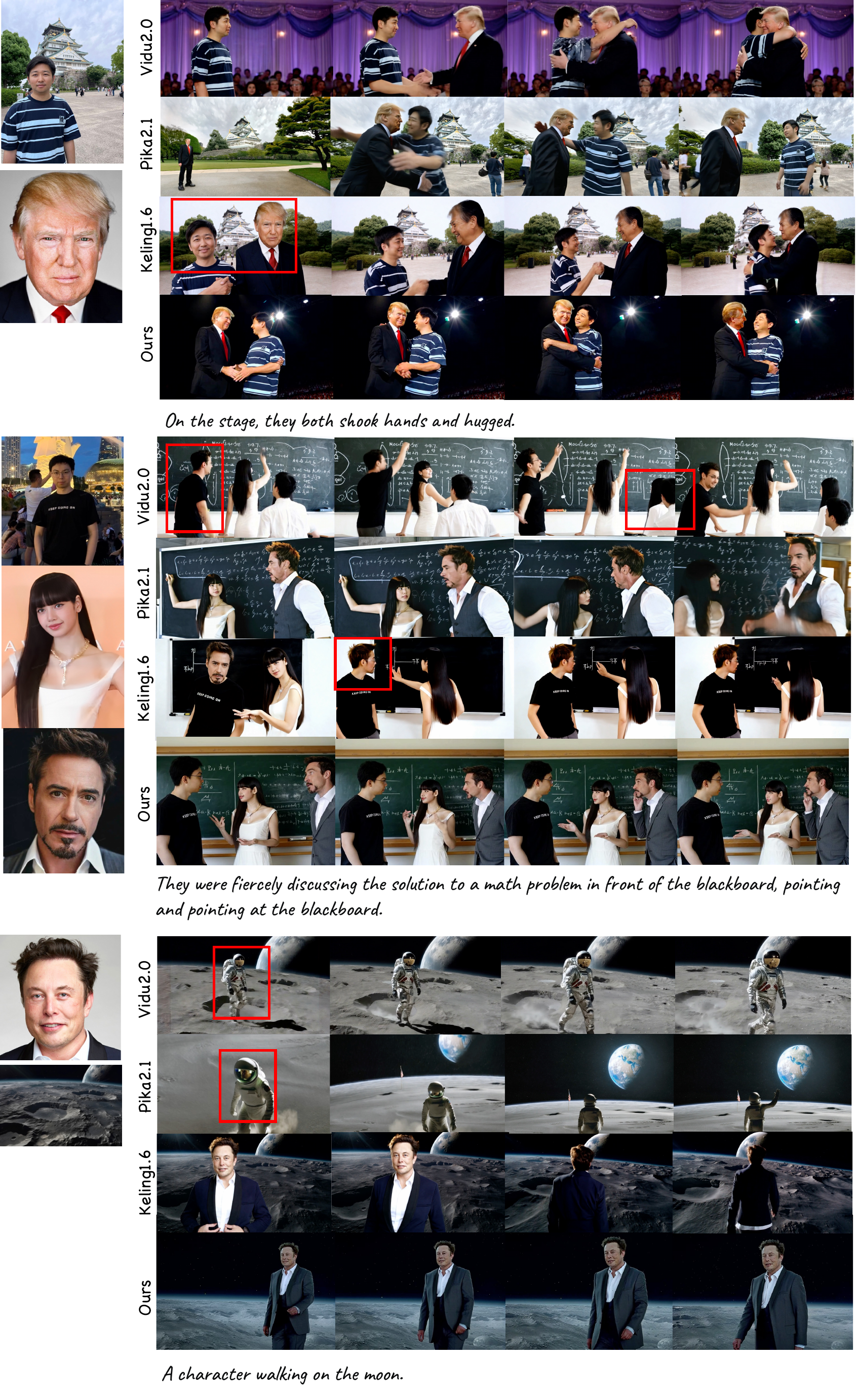} 
	\caption{Comparative results of multi-reference subject-to-video generation.}
	\label{fig:supp_mip}
\end{figure*}

Figure \ref{fig:supp_mip} displays some qualitative comparisons of multi-subject consistency generation. Firstly, Kling still reflects the I2V pattern, appearing unnatural transitions in the first few frames of the video. Additionally, in the second example with three reference images of persons, confusion issues are evident in all methods except ours. Vidu shows the first man's clothes and the second man's face, and includes a person unrelated to the reference images. Pika misses one person, and Kling also lacks one person and shows the same issue as Vidu. The final case demonstrates that Vidu and Pika appear more realistic, indicating that their text responsiveness is stronger than their subject consistency.

\subsection*{8. Limitations and future work}
\label{sec:supp:limitations}
\textbf{Limitations.} While \textit{Phantom} demonstrates strong performance in subject-consistent video generation, several challenges persist. First, handling uncommon subjects (e.g., rare animals or niche objects) remains difficult due to biases in training data coverage. Second, complex multi-subject interactions (e.g., overlapping movements or fine-grained spatial relationships) often lead to partial confusion or inconsistent relative subject sizes. Third, generating videos that strictly adhere to intricate text responses (e.g., precise spatial layouts or nuanced temporal dynamics) is limited by the current cross-modal alignment mechanism. These issues stem from three core factors: (1) gaps in dataset diversity, particularly for non-human-centric scenarios; (2) the inherent rigidity of the reference image injection strategy, which struggles to disentangle entangled features from multiple subjects; and (3) biases inherited from pre-trained base models and visual encoders, such as CLIP’s semantic oversimplification and VAE’s over-referenced details. 

\noindent \textbf{Future work.} Addressing these limitations will require multi-faceted innovations, and we propose the following directions: 

\begin{itemize}
    \item Enhanced Cross-Modal Alignment: Develop adaptive injection mechanisms that dynamically prioritize text or image conditions based on task requirements, reducing content leakage and improving text responsiveness. 
    
    \item Spatiotemporal Disentanglement: Integrate spatial-aware attention modules and physics-inspired motion priors to better model multi-subject interactions and enforce consistent relative scales. 
    
    \item Bias-Aware Training: Mitigate dataset and model biases through adversarial debiasing techniques and synthetic data augmentation for underrepresented subjects. 
    
    \item Granular Control: Explore auxiliary control signals (e.g., depth maps, segmentation masks) to complement text prompts, enabling precise alignment with complex instructions. 
    
    \item Foundation Model Adaptation: Fine-tune pre-trained encoders on domain-specific data (e.g., medical imaging, animation) to broaden \textit{Phantom}’s applicability while preserving generalization. 
\end{itemize}

By advancing these areas, \textit{Phantom} could evolve into a versatile tool for industrial applications such as virtual try-ons, interactive storytelling, and educational content creation, ultimately narrowing the gap between academic research and real-world demands.

\end{document}